\documentclass[10pt,twocolumn,letterpaper]{article}
\pdfoutput=1

\usepackage{cvpr}              

\usepackage{graphicx}
\usepackage{amsmath}
\usepackage{amssymb}
\usepackage{booktabs}
\usepackage{multirow}
\usepackage{comment}
\usepackage{xcolor}
\usepackage{soul}
\usepackage[accsupp]{axessibility} %
\usepackage{hyperref}
\hypersetup{breaklinks,colorlinks}

\usepackage[capitalize]{cleveref}
\crefname{section}{Sec.}{Secs.}
\Crefname{section}{Section}{Sections}
\Crefname{table}{Table}{Tables}
\crefname{table}{Tab.}{Tabs.}

\usepackage{xcolor}

\def\cvprPaperID{9} 
\def\confName{CVPR}
\def\confYear{2023}

\begin{document}

\title{FishEye8K: A Benchmark and Dataset for Fisheye Camera Object Detection}


\author{Munkhjargal Gochoo$^{1,2}$ 
\hspace{0.1cm} Munkh-Erdene Otgonbold$^{1,2}$ 
\hspace{0.1cm} Erkhembayar Ganbold$^{1,2}$ 
\hspace{0.1cm} Ming-Ching Chang$^{4}$\\
\hspace{0.2cm} Ping-Yang Chen$^{5}$
\hspace{0.2cm} Byambaa Dorj$^{6}$
\hspace{0.3cm} Hamad Al Jassmi$^{1,2}$
\hspace{0.3cm} Ganzorig Batnasan$^{1}$ \\
\hspace{0.3cm} Fady Alnajjar$^{1}$ 
\hspace{0.3cm} Mohammed Abduljabbar$^{1}$
\hspace{0.3cm} Fang-Pang Lin$^{7}$ 
\hspace{0.3cm} Jun-Wei Hsieh$^{3}$\\
\\
$^{1}$ Department of Computer Science and Software Engineering, United Arab Emirates University, UAE\\
$^{2}$ Emirates Center for Mobility Research, United Arab Emirates University, UAE\\
$^{3}$ College of AI and Green Energy, National Yang Ming Chiao Tung University, Taiwan\\
$^{4}$ University at Albany --- State University of New York, NY, USA\\
$^{5}$ Department of Computer Science, National Yang Ming Chiao Tung University, Taiwan\\
$^{6}$ Mongolian University of Science and Technology, Mongolia\\
$^{7}$ National Center for High-Performance Computing, Taiwan\\
{\small 
\url{mgochoo@uaeu.ac.ae}, \hspace{1mm}
\url{omunkuush@uaeu.ac.ae}, \hspace{1mm} 
\url{eganbold@uaeu.ac.ae}, \hspace{1mm} 
\url{mchang2@albany.edu}, \vspace{-1mm}}\\
{\small
\url{pingyang.cs08@nycu.edu.tw}, \hspace{1mm}
\url{dorj@must.edu.mn}, \hspace{1mm}
\url{h.aljasmi@uaeu.ac.ae}, \hspace{1mm} 
\url{fady.alnajjar@uaeu.ac.ae},} \vspace{-1mm}\\
{\small 
\url{gbatnasan@uaeu.ac.ae}, \hspace{1mm}
\url{201970087@uaeu.ac.ae}, \hspace{1mm}
\url{fplin@narlabs.org.tw}, \hspace{1mm}
\url{jwhsieh@nctu.edu.tw}} \vspace{-4mm} \\ 
}

\maketitle

\begin{abstract}
\vspace{-3mm}
With the advance of AI, road object detection has been a prominent topic in computer vision, mostly using perspective cameras. Fisheye lens provides omnidirectional wide coverage for using fewer cameras to monitor road intersections, however with view distortions. To our knowledge, there is no existing open dataset prepared for traffic surveillance on fisheye cameras.
This paper introduces an open FishEye8K benchmark dataset for road object detection tasks, which comprises 157K bounding boxes across five classes (Pedestrian, Bike, Car, Bus, and Truck). In addition, we present benchmark results of State-of-The-Art (SoTA) models, including variations of YOLOv5, YOLOR, YOLO7, and YOLOv8. The dataset comprises 8,000 images recorded in 22 videos using 18 fisheye cameras for traffic monitoring in Hsinchu, Taiwan, at resolutions of 1080$\times$1080 and 1280$\times$1280.
The data annotation and validation process were arduous and time-consuming, due to the ultra-wide panoramic and hemispherical fisheye camera images with large distortion and numerous road participants, particularly people riding scooters. To avoid bias, frames from a particular camera were assigned to either the training or test sets, maintaining a ratio of about 70:30 for both the number of images and bounding boxes in each class. Experimental results show that YOLOv8 and YOLOR outperform on input sizes 640$\times$640 and 1280$\times$1280, respectively. The dataset will be available on the GitHub \href{https://github.com/MoyoG/FishEye8K}{link} with PASCAL VOC, MS COCO, and YOLO annotation formats. The FishEye8K benchmark will provide significant contributions to the fisheye video analytics and smart city applications.
\vspace{-3mm}

\end{abstract}

\vspace{-8mm}
\section{Introduction}
\label{sec:intro}

\begin{figure}[t]
\centerline{
   \includegraphics[width=1\linewidth]{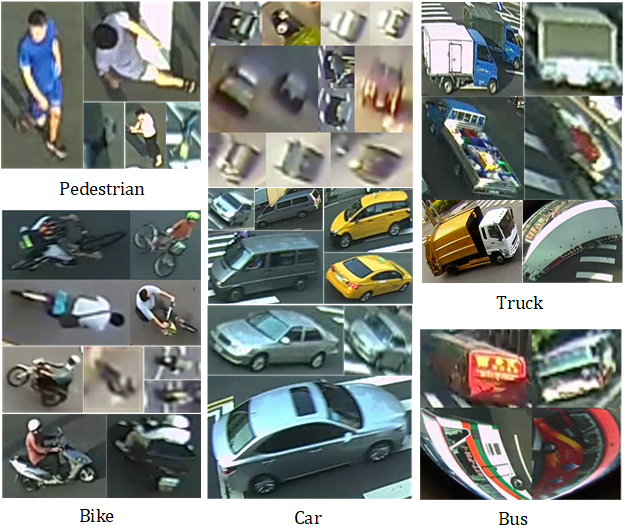}
\vspace{-2mm}
}   
\caption{Sample of the 5 classes in the FishEye8K dataset: 
Pedestrian (all visible people on the streets),
Bike (people riding bicycles, motorcycles, or
scooters), 
Car (light vehicles such as sedans, SUVs, Vans,
{\em etc.}), 
Bus, and Truck (dump-truck, semi-trailers, {\em etc.})
}
\label{fig:Classes}
\vspace{-5mm}
\end{figure}


\begin{table*}[t]
\centerline{
\footnotesize
\renewcommand{\tabcolsep}{2pt}
\begin{tabular}{||c c c c c c c c c c c c c||}
\hline
\textbf{Dataset} & \textbf{Frame} & \textbf{Boxes} & \textbf{Task} & \textbf{Vehicles} & \textbf{Pedestrian} & \textbf{Weather} & \textbf{Occlusion} & \textbf{Altitude} & \textbf{View} & \textbf{Classes} & \textbf{Location} & \textbf{Type} \\ [1ex]
\hline\hline  
MIT-Car 2000\cite{MIT-Car} & 1.1K & 1.1K & D & + & & & & & & - & Surveillance & 2D \\
 KITTI-D 2014\cite{KITTI-D} & 15K & 80.3K & D & + & + &  & + & & & 3 & Car & 2D \\
 UA-DETRAC 2015\cite{UA-DETRAC} & 140K & 1210K & D,T & + & & + & + & & & 4 & Surveillance & 2D\\
 Detection in LLC 2017\cite{Detection_in_LLC} & 7.5K & 15K & D & + & & + & & & & 12 & Car & 2D\\
  CARPK 2017\cite{CARPK} & 1.5K & 90K & D & + & & & & & & - & Drone & 2D \\
 UAVDT 2017\cite{UAVDT} & 80K & 841.5K & D,T & + & & + & + & + & + & - & Drone & 2D\\
 NEXET 2017\cite{NEXET} & 50K & - & D & + & & + & & & & 5 & Car & 2D\\
 BDD100k 2018\cite{BDD100K} & 5.7K & - & D,T & + & + & + & & & & 10 & Car & 2D\\
 AAU RainSnow 2018\cite{AAURainSnow} & 2.2K & 13297 & D,Seg & + & & + & & & &  & Surveillance & RGB\&Thermal \\
 MIO-TCD CCTV 2018\cite{MIO-TCD} & 113K & 200K & D & + & & + & & & & 5 & Surveillance & 2D\\
 BDD100k Adas 2018\cite{yu2020bdd100k} & 100K & 250K & D,Seg & + &  & + & & & & 10 & Car & 2D\\
 Woodscape 2018/2019\cite{woodscape} & 10K & - & D,3D,T & + & & + & & & & 7 & Car	& Fish-Eye\\
CityFlow2D 2021\cite{AI-City} & - & 313.9K & D,T & + & & & & & & & Surveillance & 2D\\
FishEye8K 2023 [\textbf{our}] & 8K & 157.0K & D & + & + &  & & & + & 5 & Surveillance & Fish-Eye\\ [0.5ex]
\hline
\end{tabular}
\vspace{-2mm}
}  
\caption{Summary of existing road traffic datasets. The second and third columns $(1K=10^3)$ indicate the number of images containing at least one object on them and the unique object bounding boxes. Remaining columns: additional attributes for each dataset, i.e., $"D"$: target is a detection task, $"3D"$: target is a three-dimensional detection task, $"T"$: target is a tracking task, and the $"Seg"$: target is a segmentation task. }
\label{table:1}
\vspace{-4mm}  
\end{table*}

Fisheye lenses have gained popularity owing to their natural, wide, and omnidirectional coverage, which traditional cameras with narrow fields of view (FoV) cannot achieve. In traffic monitoring systems, fisheye cameras are advantageous as they effectively reduce the number of cameras required to cover broader views of streets and intersections. Despite these benefits, fisheye cameras present distorted views that necessitate a non-trivial design for image undistortion and unwarping or a dedicated design for handling distortions during processing. It is worth noting that, to the best of our knowledge, there is no open dataset available for fisheye road object detection for traffic surveillance applications. The WoodScape dataset~\cite{woodscape} was collected using an in-car fisheye dash camera; however, it was intended for self-driving scenarios.


In this paper, we present a new open {\bf FishEye8K} benchmark dataset for the training and evaluation of 2D road object detection tasks. The FighEye8K dataset consists of 8,000 image frames with 157K bounding box annotations of 5 object classes, namely, Pedestrian, Bike, Car, Truck, Bus, and Truck; see Figure~\ref{fig:Classes}. 
A total of 22 short (8 to 20 minutes) videos were extracted from many hour-long videos collected from 35 fisheye cameras.
These traffic surveillance cameras are properties of the police department of Hsinchu City, Taiwan, and our data collection is free from user consent agreements or license issues. 
However, efforts are performed in blurring out visible faces and license plates in the video frames. 
The dataset comprises different traffic patterns and conditions, including urban highways, road intersections, various illumination, and shooting angles of the five road object classes in various scales. 

The labeling of objects of interest is meticulous. Specifically, we labeled all visible and recognizable objects even if they are located far away. The FishEye8K sample images are split into the training and test sets, with a ratio of about 70:30. Efforts are made to keep a similar ratio for each class of road objects. To avoid bias, the train and test sets do not share frames from the same camera. Annotations are provided in several standard formats, including Pascal-VOC\cite{Pascal-Voc}, MS COCO~\cite{COCO}, and YOLO~\cite{YOLO}.

We also provide benchmarking results of the latest State-of-The-Art (SoTA) two-stage object detection models, including YOLOv5x~\cite{YOLOv5}, YOLOR~\cite{YOLOR}, YOLOv7~\cite{wang2022yolov7}, and YOLOv8, and report in standard metrics including $Precision$, $Recall$, $mAP$s, $AP\textsubscript{S}$, $AP\textsubscript{M}$, $AP\textsubscript{L}$, $F1-score$, and their inference time.


The FishEye8K benchmark dataset will be available at \url{https://github.com/MoyoG/FishEye8K} upon paper acceptance.

\begin{figure*}[t]
\centerline{
   \includegraphics[width=1\linewidth]{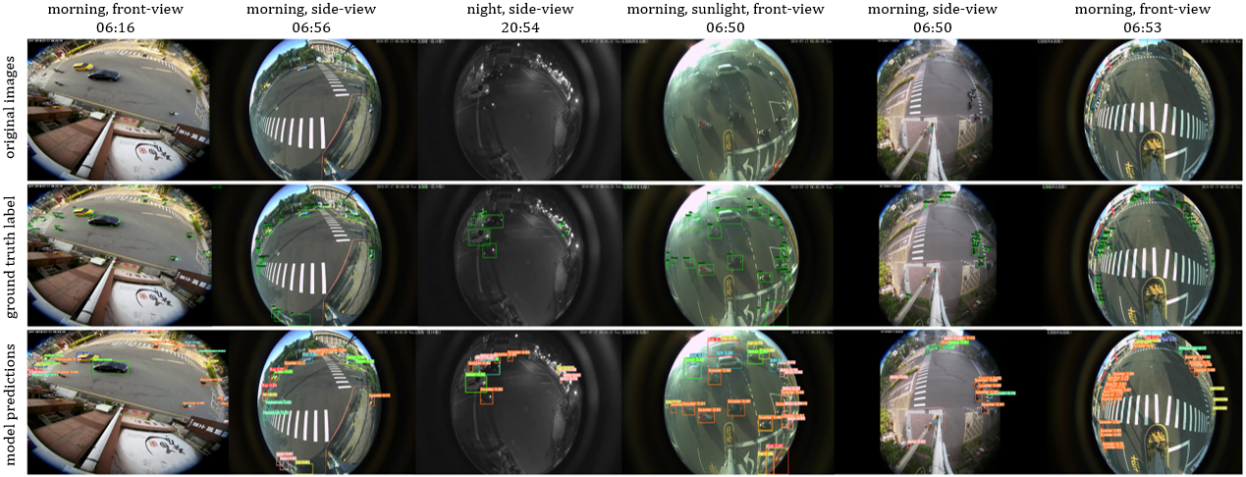}
\vspace{-2mm}   
}   
\caption{Sample images of FishEye8K dataset:
(Top) the original unlabelled images,
(Middle) the labeled ground truths,
(Bottom) the YOLOv5x6~\cite{YOLOv5} detected objects. 
The columns illustrate several viewing angles, time of day, various intersections and road participants in the dataset. 
}   
\label{fig:fisheye10ksample3}
\vspace{-4mm}
\end{figure*}

\section{Related Works}
\label{sec:Related works}

\textbf{Road datasets.} High-resolution, diverse, and large-scale road datasets play a critical role in advancing and enhancing traffic monitoring systems. In the last decade, the number of open road datasets~\cite{MIT-Car, KITTI-D, UA-DETRAC, Detection_in_LLC, CARPK, UAVDT, NEXET, BDD100K, AAURainSnow, MIO-TCD, yu2020bdd100k, woodscape, AI-City} for 2D and 3D road object detection, single and multiple object tracking, object segmentation tasks have significantly increased. 
Table~\ref{table:1} provides a summary of popular road datasets that are used in both model development as well as for benchmarking and public contests. 
In terms of camera locations, the following datasets are captured using fixed surveillance cameras: MIT-Car~\cite{MIT-Car}, UA-DETRAC~\cite{UA-DETRAC}, AAU RainSnow~\cite{AAURainSnow}, MIO-TCD~\cite{MIO-TCD}, and AI-City~\cite{AI-City} datasets.
The CARPK~\cite{CARPK} and UAVDT~\cite{UAVDT} dataasets are captured using drones.
The KITTI~\cite{KITTI-D}, Detection in LLC~\cite{Detection_in_LLC}, NEXET~\cite{NEXET}, BDD100K~\cite{BDD100K}, and Woodscape~\cite{woodscape} datasets are captured using in-dash cameras mounted on a car. 
In terms of FoV, all the datasets were constructed using standard perspective cameras, with the drawback of narrow FoV.
The only exception is the WoodScape dataset~\cite{woodscape} that are captured using an in-dash 180\textdegree{} fisheye camera. 
To our knowledge, the proposed FishEye8K dataset is the first of the kind among the open datasets, that are designed and constructed specifically for the development and evaluation of road object detection using fisheye traffic surveillance cameras.

\textbf{Fixed perspective traffic camera-based datasets.} 
Table~\ref{table:1} shows that most datasets are captured using fixed, perspective cameras, which are limited by the narrow FoV. All the datasets have annotations for 2D road object detection task; on top of it, a few datasets \cite{UAVDT,AI-City} have multiple objects tracking annotation, and one \cite{AAURainSnow} has segmentation mask annotation.
In 2000, MIT-Car dataset \cite{MIT-Car} was publicly offered as a flagship dataset pioneering the road automation research field. The dataset has 1.1K frames, including 1.1K bounding boxes for the vehicle detection task. In 2016, UA-DETRAC \cite{UA-DETRAC} dataset was offered with 140K frames, including rich annotations of illumination, vehicle type, occlusion, and 1210K bounding boxes. The dataset has four classes (car, van, bus, and others) for detection and multiple object detection tasks. In the same year, similarly, MIO-TCD CCTV \cite{MIO-TCD} dataset is offered with 113K frames, including 200K bounding boxes for the detection task. In 2018, the AAU RainSnow \cite{AAURainSnow} dataset was offered as a benchmark for evaluating state-of-the-art rain removal algorithms. The dataset has 22 five-minute real-world camera video sequences collected from 7 urban intersections covering various weather conditions, i.e., snow, rain, haze, and fog. They have extracted 100 frames from each five-minute video to construct 2200 frames, including 13297 bounding boxes. Recently, in 2021, AI-City Challenge \cite{AI-City} was held, including vehicle detection and re-identification on CityFlowV2-ReID dataset and multi-target multi-camera vehicle tracking challenge on CityFlow2D dataset. CityFlow2D dataset has 313.9K bounding boxes for 880 distinct vehicles.

\textbf{Drone based datasets.}
Lately, drone road datasets have been publicly offered in the literature, namely CARPK \cite{CARPK} and UAVDT \cite{UAVDT}. Both datasets were captured from a high altitude with a viewing angle of the top by narrow FOV cameras for the drone-based road monitoring systems. Thus they are not suitable for fixed surveillance camera-based traffic monitoring.

\begin{figure*}[t]
\centerline{
    \includegraphics[width=1\linewidth]{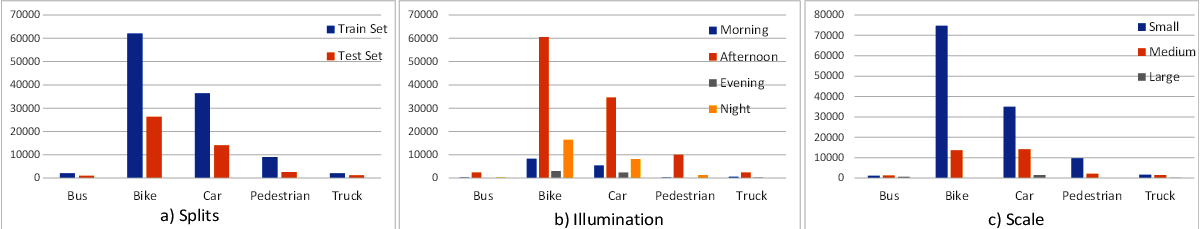}
\vspace{-2mm}
}
\caption{The class distributions of objects in terms of (a) Splits for FishEye8K dataset; (b) Illumination; and (c) Scale.}
\label{fig:Diverse}
\vspace{-4mm}
\end{figure*}


\section{The FishEye8K Dataset}
\label{3}

We provide detailed information on the new FishEye8K road object detection dataset. The dataset consists of 8,000 annotated images with 157K bounding boxes of five object classes. Figure~\ref{fig:fisheye10ksample3} shows sample images of the wide-angle fisheye views, which provide new opportunities for large coverage, but also new challenges of large distortions of the road objects.



\subsection{Video Acquisition}
\label{3:1}

We have acquired a total of 35 fisheye videos captured using 20 traffic surveillance cameras at 60 FPS in Hsinchu City, Taiwan.
Among them, the first set of 30 videos ({\bf Set 1}) was recorded by the cameras mounted at Nanching Hwy Road on July 17, 2018, with $1920 \times 1080$ resolution, and each video lasts about 50-60 minutes. 
The second set of 5 videos ({\bf Set 2}) was recorded at $1920 \times 1920$ resolution, and each video lasts about 20 minutes. 

All cameras are the property of the local police department, so there is no issue of user consent or license issues. All images in the dataset will be made available to the public for academic and R\&D use. 


\subsection{Dataset Preparation and Characteristics}
\label{3:2} 

\textbf{Sampling.} We chose 18 videos from the recorded footage, with 15 videos coming from Set 1. These were cropped into shorter videos, each lasting approximately 8 to 10 minutes, except for one that lasted 16 minutes. Using a sampling method of one frame per 50 and 200 frames for Set 1 and Set 2 videos, respectively, we extracted over 10,000 frames. The resulting images were then resized to $1080 \times 1080$ and $1280 \times 1280$ for Set 1 and Set 2, respectively.


To incorporate a wide range of perspectives on road conditions, we carefully selected videos for our dataset that feature diverse camera angles, including side-view and front-view shots, as well as varying video quality. The dataset also includes images from different intersection types, such as T-junctions, Y-junctions, cross-intersections, midblocks, pedestrian crossings, and non-conventional intersections. The videos were captured under various lighting conditions, including morning, afternoon, evening, and night, and diverse traffic congestion levels ranging from free-flowing to steady and busy. Figure~\ref{fig:fisheye10ksample3} illustrates some of the wide-ranging road conditions with ground truth annotations of road objects and detection results obtained from YOLOv5x6~\cite{YOLOv5}.


\textbf{Object classes:} We annotate 5 major classes for road objects, namely, \textbf{Pedestrian} (all visible people on the streets), \textbf{Bike} (riders on bicycles, motorcycles, or scooters), \textbf{Car} (light vehicles such as sedans, SUVs, vans, {\em etc.}), \textbf{Bus}, and \textbf{Truck} (dump-truck, semi-trailers, {\em etc.}). 

\textbf{Distant objects:} 
The wide fisheye lens creates a wide FoV but also results in a panoramic hemispherical image that is notably distorted with a barrel effect.
Additionally, the camera has a tendency to produce blurred images of objects located around the edges of the lens. As a consequence, distant objects can appear minuscule and indistinct. Annotating these distant objects can be an arduous or even impossible task due to their lack of clarity.


\textbf{Illumination:} 
Four categories of illumination conditions were identified, namely morning (sunrise), afternoon (sunny), evening (sunset), and night. The distribution of video sequences based on their respective illumination attributes is illustrated in Figure~\ref{fig:Diverse}(b), with the majority of bounding boxes falling under the afternoon category. Night-time sequences follow in second place, with morning and evening categories trailing behind respectively. Notably, the distribution of classes across all times of day is remarkably similar


\textbf{Object scale:} We define the scale of the bounding boxes of road participants based on their size (length and width) in pixels. The MS COCO evaluator is employed for small and medium, and large scaled objects. However, as the size of the image grows toward $1080\times1080$ or $1280\times1280$, respectively for Sets 1 and 2, we doubled the size of standard scales, i.e., \emph{small} (pixels $\leq$ 64$\times$64), \emph{medium} (64$\times$64 $<$ pixels $\leq$ 192$\times$192), and \emph{large} (pixels $>$ 192$\times$192). The distribution of road participants in the dataset in terms of scale is presented in Figure~\ref{fig:Diverse} (c), where small and medium-scaled objects make the most of the dataset. Bus and Truck classes have a similar number of small and medium scaled objects. On the contrary, other classes have a comparatively high number of small-scaled objects than medium and large-scale objects.

\begin{figure*}[t]
\centerline{
   \includegraphics[width=0.9\linewidth]{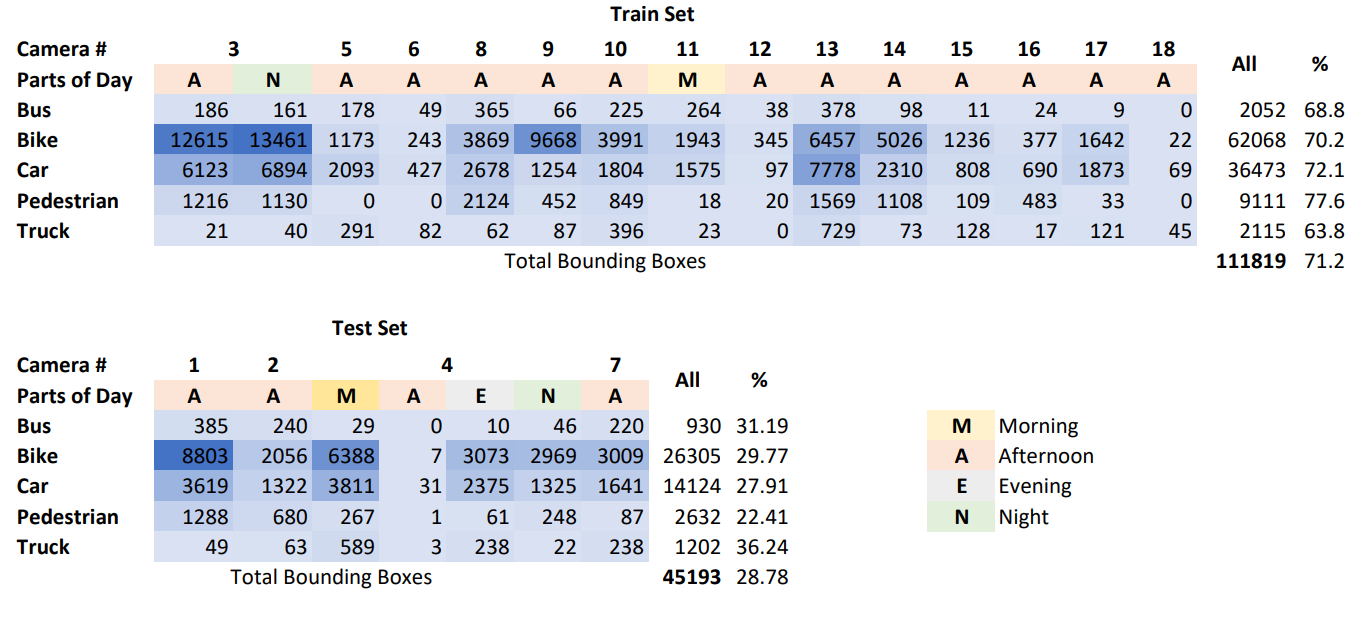}
\vspace{-4mm}
} 
\caption{Heat maps represent the number of extracted objects per class from all 22 short videos recorded by 18 cameras for training and test sets of the FishEye8K dataset. For the training set, the darkest blue color refers to 13461 labeled bikes from the video recorded at night with Camera 3.}
\label{fig:TrainTestSplit}
\vspace{-4mm}
\end{figure*}

\subsection{Annotation}
\label{3:3}

\textbf{Annotation Rule.} The road participants were annotated based on their clarity and recognizability to the annotators, regardless of their location. In some cases, distant objects were also annotated based on this criterion. 

\textbf{Annotation.} Two researchers/annotators manually labeled over 10,000 frames using the DarkLabel annotation program over a period of one year. After cleaning the dataset, a total of 8,000 frames containing 157012 bounding boxes remained. Unsuitable frames were removed, including those featuring road participants outside the five classes of interest.

The distribution of objects per class for each video is depicted in Figure~\ref{fig:TrainTestSplit}. Notably, the night video captured by Camera 3 has the highest number of objects. In this dataset, the dominant classes are Bike (88,373) and Car (50,597), which can be attributed to the semi-tropical location of the country where the videos were recorded. On the other hand, the classes of Truck (3,317) and Bus (2,982) have the lowest number of objects, rendering the dataset highly imbalanced. Figure~\ref{fig:Classes} displays a selection of samples from all classes, showcasing various scales. Furthermore, the distributions of classes are depicted as bar graphs in Figure~\ref{fig:Diverse}.

 For the sake of convenience, we provide three different formats for the annotations of FishEye8K datasets, i.e., Pascal-VOC\cite{Pascal-Voc}, MS COCO\cite{COCO}, and YOLO\cite{YOLO}.

\subsection{Validation}
\label{3:4}

Given the complexity and effort required for the labeling task, human errors were inevitable, and it was necessary to correct them to avoid inaccurate results. Therefore, in order to minimize human error, we employed two semi-automatic approaches to validate all bounding boxes.

In the case of mislabeled objects, we followed a two-step approach. Firstly, we cropped and copied the objects based on their respective bounding boxes into the corresponding directories. Secondly, our annotators manually verified if the objects were correctly placed in their designated directories through simple inspection, which is highly accurate and requires less time and effort. However, this approach is blind to objects that were not labeled in the first place, which is known as a missing label error. To address this issue, we inspected the False Positives generated by the YOLOv7 model \cite{wang2022yolov7} trained on FishEye8K, which helped identify numerous missing label errors. This approach was especially effective in identifying errors in distant areas and regions with high traffic density of vehicles and bikes.

\subsection{Dataset Splits}
\label{3:5}
In order to minimize dataset bias, we ensured that frames from the same camera were not included in both the train and test sets. Specifically, all frames from a given camera were assigned to either the train or test set. Figure~\ref{fig:TrainTestSplit} illustrates the heat maps of 22 videos (captured during morning, afternoon, evening, and night) recorded by Cameras 1-18, from which all images were extracted to create the FishEye8K dataset. To satisfy the criteria, we selected Cameras 1, 2, 4, and 7 for the test set and the remaining cameras for the training set. This division resulted in a training set that constitutes 66.07\% of the dataset, while the test set constitutes 33.93\%.

In order to maintain a roughly 70:30 ratio of objects for each class, the training set was composed of 111,835 objects and the test set contained 45,193 objects, which correspond to 71.28\% and 28.78\% of all objects, respectively. The classes Bike, Bus, and Car follow this ratio in both sets.

\subsection{Data Anonymization}
\label{3:6}

The identification of road participants such as people's faces and vehicle license plates from the dataset images was found to be unfeasible due for various reasons. The cameras used for capturing the images were installed at a higher ground level, making it difficult to capture clear facial features or license plates, especially when they are far away. Additionally, the pedestrians are not looking at the cameras, and license plates appear too small when viewed from a distance. However, to maintain ethical compliance and protect the privacy of the road participants, we blurred the areas of the images containing the faces of pedestrians and the license plates of vehicles, whenever they were visible.

\section{Benchmark}

\subsection{One-Stage 2D Object Detection Methods}
\label{4:1}

In order to assess the performance of 2D object detection methods, particularly for pedestrian and vehicle detection, we conducted a benchmark of the latest state-of-the-art one-stage detectors. Our selection process involved reviewing the literature and identifying the best-performing models, including YOLOv5\cite{YOLOv5}, YOLOR\cite{YOLOR}, YOLOv7\cite{wang2022yolov7}, and the latest YOLOv8. One-stage detectors predict bounding boxes on images without requiring a region proposal step, which results in faster processing times and makes them suitable for real-time applications. However, these detectors prioritize inference speed and may not perform as well for recognizing irregularly shaped objects or groups of small objects. Table~\ref{table:2} presents the results of our benchmark of the one-stage detectors.

\begin{table*}[t]
\footnotesize
\centerline{
\renewcommand{\tabcolsep}{5pt}
\begin{tabular}{|c|l|c|l|l|l|l|l|l|l|l|c|}
\hline
\multicolumn{1}{|l|}{\textbf{Model}}         & \textbf{Version} & \textbf{Input Size} & \textit{\textbf{Precision}} & \textit{\textbf{Recall}} & \textit{\textbf{mAP\textsubscript{0.5}}} & \textit{\textbf{mAP\textsubscript{.5-.95}}} & \textit{\textbf{F1-score}}         & \textit{AP\textsubscript{S}}    & \textit{AP\textsubscript{M}}    & \textit{AP\textsubscript{L}}    & \multicolumn{1}{c|}{\textbf{\begin{tabular}[c]{@{}c@{}}Inference\\ {[}ms{]}\end{tabular}}} \\ \hline
\multirow{2}{*}{YOLOv5 \cite{YOLOv5}}                      & YOLOv5l6         & 1280$\times$1280                & 0.7929                      & 0.4076                   & 0.6139                   & 0.4098                      & 0.535                       & 0.1299          & 0.434           & 0.6665          & 22.7                                                                                       \\ \cline{2-12} 
                                             & YOLOv5x6         & 1280$\times$1280                 & \textbf{0.8224}             & 0.4313                   & 0.6387                   & 0.4268                      & 0.5588                      & 0.133           & 0.452           & 0.6925          & 43.9                                                                                       \\ \hline
\multicolumn{1}{|l|}{\multirow{2}{*}{YOLOR \cite{YOLOR}}} & YOLOR-W6         & 1280$\times$1280                 & 0.7871                      & 0.4718                   & 0.6466                   & \textbf{0.4442}             & 0.5899 & 0.1325          & 0.4707          & 0.6901          & 16.4                                                                                       \\ \cline{2-12} 
\multicolumn{1}{|l|}{}                       & YOLOR-P6         & 1280$\times$1280                 & 0.8019                      & 0.4937                   & \textbf{0.6632}          & 0.4406                      &   0.6111 & 0.1419          & 0.4805          & \textbf{0.7216} & \textbf{13.4}                                                                              \\ \hline
\multirow{2}{*}{YOLOv7 \cite{wang2022yolov7}}                      & YOLOv7-D6        & 1280$\times$1280                 & 0.7803                      & 0.4111                   & 0.3977                   & 0.2633                      & 0.5197                      & 0.1261          & 0.4462          & 0.6777          & 26.4                                                                                       \\ \cline{2-12} 
                                             & YOLOv7-E6E       & 1280$\times$1280                 & 0.8005                      & \textbf{0.5252}          & 0.5081                   & 0.3265                      & \textbf{0.6294}             & \textbf{0.1684} & \textbf{0.5019} & 0.6927          & 29.8                                                                                       \\ \hline \hline
\multirow{2}{*}{YOLOv7 \cite{wang2022yolov7}}                      & YOLOv7           & 640$\times$640                & 0.7917                      & 0.4373                   & 0.4235                   & 0.2473                      & 0.5453                      & 0.1108          & 0.4438          & 0.6804          & \textbf{4.3}                                                                               \\ \cline{2-12} 
                                             & YOLOv7-X         & 640$\times$640                 & 0.7402                      & \textbf{0.4888}          & 0.4674                   & 0.2919                      & \textbf{0.5794}             & \textbf{0.1332}          & \textbf{0.4605}          & \textbf{0.7212}          & 6.7                                                                                        \\ \hline
\multirow{2}{*}{YOLOv8}                      & YOLOv8l          & 640$\times$640                  & 0.7835                      & 0.3877                   & 0.612                    & 0.4012                      &              0.5187 & 0.1038          & 0.4043          & 0.6577          & 8.5                                                                                        \\ \cline{2-12} 
                                             & YOLOv8x          & 640$\times$640                  & \textbf{0.8418}             & 0.3665                   & \textbf{0.6146}          & \textbf{0.4029}                      & 0.5106 & 0.0997          & 0.4147          & 0.7083          & 13.4                                                                                       \\ \hline
\end{tabular}
\vspace{-2mm}
}
\caption{Results of state-of-the-art models trained on FishEye8K datasets. The table consists of two groups of various versions of YOLO object detection models for input sizes 1280$\times$1280 and 640$\times$640.}
\label{table:2}
\vspace{-4mm}
\end{table*}

\subsection{Training Procedure}
\label{4:2}

We utilized several frameworks and platforms, i.e., Darknet\cite{darknet}, Pytorch\cite{pytorch}, and PaddlePaddle\cite{paddle}, for the model training.

\textbf{Hyperparameters.} All YOLO variations were pre-trained on MS COCO \cite{COCO} dataset. Among the models, we trained four models (YOLOv7 \cite{wang2022yolov7}, YOLOv7-X\cite{wang2022yolov7}, YOLOv8l, and YOLOv8x on the input size 640$\times$640. Six models (YOLOv5x6 \cite{YOLOv5}, YOLOv5l6 \cite{YOLOv5}, YOLOR-W6 \cite{YOLOR}, YOLOR-P6 \cite{YOLOR}, YOLOv7-D6 \cite{wang2022yolov7}, YOLOv7-E6E \cite{wang2022yolov7}) on the input size 1280$\times$1280. All models have trained with the same training procedures for 250 epochs, Adam \cite{ADAM} optimizer with the momentum of 0.937 except for YOLOv5 \cite{YOLOv5} which employed SGD optimizer. The confidence and NMS (Non Max Suppression) IoU (Intersection over Union) thresholds were both 0.5, and a learning rate of 0.01.

\textbf{Data preprocessing.} For the purpose of training and testing, the input images were resized to 640$\times$640 and 1280$\times$1280 for particular models, see Table~\ref{table:2}.

\textbf{Loss Objective.} We employed the Focal loss\cite{Focal} as it is commonly used in the multi-object detection and multi-label image classification domain. The loss function is defined as: 
\begin{equation}
FL(p_{t}) = -\alpha_{t}(1-p_{t})^\gamma log(p_{t}),
\label{eq:1}
\end{equation}
where by default $\gamma = 0.5$ and $\alpha = 0.5$, $p_{t}$ is the predicted probability for the object indexed by $t$.

\subsection{Metrics}
\label{4:3}

All models are analyzed and evaluated with the same metrics, i.e., $Precision$, $Recall$, $mAP$s, $AP\textsubscript{S}$, $AP\textsubscript{M}$, $AP\textsubscript{L}$, $F1-score$, and their inference time.

\textbf{\emph{F1-score}} metric measures the balance between \emph{Precision} and \emph{Recall}. When both \emph{Precision} and \emph{Recall} are high, the $F1$ score is high as well, indicating good model performance. On the other hand, a low $F1$ score indicates that the \emph{Precision} and \emph{Recall} values are imbalanced, and the model is not performing well. The $F1$ score is calculated as below:
\begin{equation}
   F_{1}=\frac{2 \times Precision \times Recall}{Precision + Recall} 
   \label{eq:3}
\end{equation}

\textbf{Average Precision ($AP$)} represents all \emph{Precision} and \emph{Recall} values into a single score. The $AP$ is calculated according to:
\begin{equation}
AP = \sum_{k=0}^{n-1} [Recall_{(k+1)}-Recall_{(k)}]*Precision_{(k+1)},
\nonumber
\label{eq:6}
\end{equation}
where $k$ is an index of the frame, and $n$ is the number of frames for a given class.


\textbf{Intersection over Union (IoU).} The model predicts the bounding boxes of the detected objects; however, it is expected that the predicted box will not match exactly the ground truth box.  Intersection over Union (IoU) is employed to quantify the measure to score how the ground truth and predicted boxes match:
$
   IoU=\frac{Intersection \; Area}{Union \; Area}.
   \label{eq:4}
$

\textbf{Normalized Confusion Matrix} is used to determine the prediction quality of the model by each class. A confusion matrix is made up of 4 components, namely, True Positive (\emph{TP}), True Negative (\emph{TN}), False Positive (\emph{FP}), and False Negative (\emph{FN}). 

\textbf{Mean Average Precision ($mAP$s)} is the mean of the $AP$s for all classes. The $mAP$ of the object detection model is calculated according to:
\begin{equation}
mAP = \frac{1}{n}\sum_{k=1}^{n} AP_{k},
\label{eq:5}
\end{equation}
where $n$ is the number of classes in the dataset and $AP(k)$ is the average precision $(AP)$ for a given class $k$.

\subsection{Performance}
\label{4:4}

In this subsection, we report the experimental results of variations of YOLOv5\cite{YOLOv5}, YOLOR\cite{YOLOR}, YOLOv7\cite{wang2022yolov7}, and YOLOv8, which are trained on DGX-1 GPU server accessed by internal web-based job and resource allocation system \cite{DGX-1}. 

Table~\ref{table:2} presents two sets of models that were trained on the FishEye8K dataset, with input sizes of 1280$\times$1280 and 640$\times$640.

\subsubsection{Results on Input Size 640 $\times$ 640}
\label{4:4:2}



For input size 640$\times$640, the highest two $mAP\textsubscript{0.5}$s of 0.6146 and 0.612 are achieved by YOLOv8x and YOLOv8l, respectively. The lowest $mAP\textsubscript{0.5}$s of 0.4235 is result of YOLOv7 \cite{wang2022yolov7}. In terms of \emph{F1-score} and \emph{Recall}, YOLOv7-X achieved the highest performance with 0.5794 and 0.4888, respectively. Further, in terms of object scale, YOLOv7-X outperformed on all three scales (small, medium, and large) as well.

\begin{figure}[t]
\centerline{
   \includegraphics[width=0.92\linewidth]{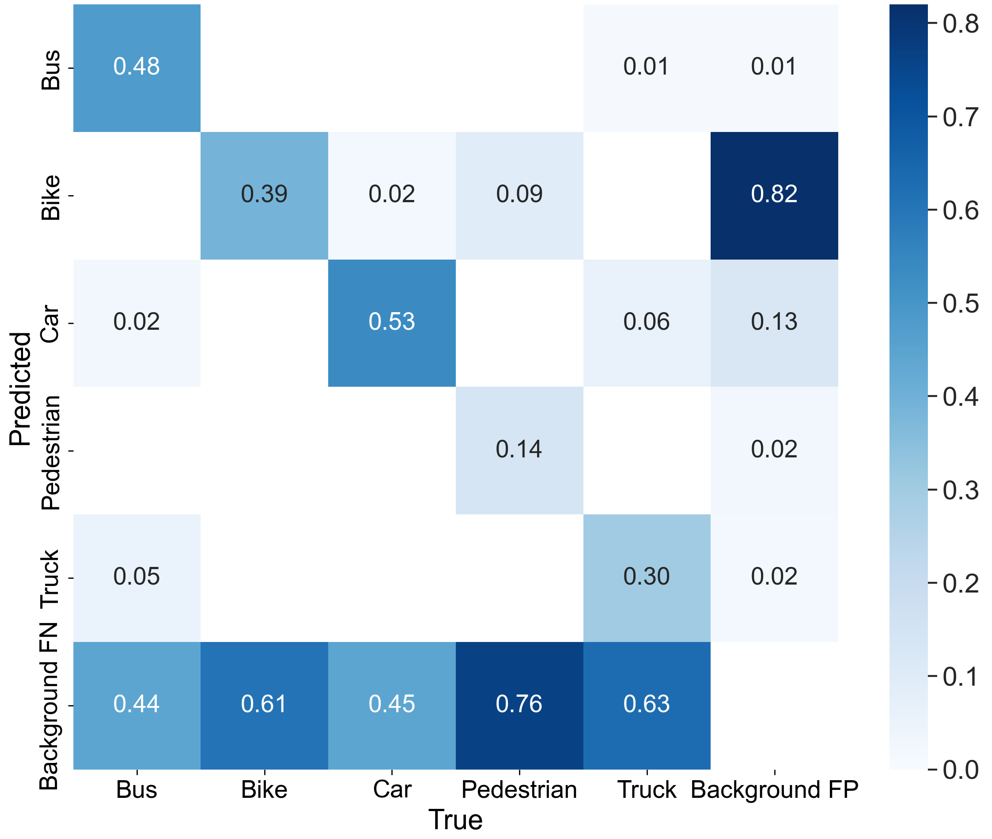}
\vspace{-2mm}
}   
\caption{Normalized Confusion Matrix of YOLOv8x model on the input size 640 $\times$ 640.} 
\label{fig:CM_yolo8}
\vspace{-4mm}
\end{figure}

The confusion matrix for the best-performing model, YOLOv8x, on the input size of 640$\times$640, is presented in Figure~\ref{fig:CM_yolo8}, and Table \ref{table:4} tabulates the results. The Car class achieved the highest $mAP\textsubscript{0.5}$ score of 0.749, followed by Bus, Bike, Truck, and finally Pedestrian with a score of 0.4596. Surprisingly, the Bike class had the highest $FP$ rate of 0.82, with many objects mispredicted as Bike on the background. Additionally, a significant portion of objects across all classes were undetected, with normalized $FN$s ranging from 0.45 to 0.84. However, the model performed significantly well in terms of \emph{Precision} for all classes, with values ranging from 0.74 to 0.94. The Pedestrian class had the lowest normalized \emph{TP} rate at 0.14, indicating incorrect predictions of this class as others, mainly as Background which has the maximum normalized \emph{FN} rate at 0.76.

\begin{table}[t]
\footnotesize
\centerline{
\begin{tabular}{|llllll|}
\hline
\multicolumn{6}{|c|}{YOLOv8x-640$\times$640}\\ \hline
\multicolumn{1}{|l|}{\textbf{Classes}} & \multicolumn{1}{l|}{\textbf{\textit{Precision}}}      & \multicolumn{1}{l|}{\textbf{\textit{Recall}}}      & \multicolumn{1}{l|}{\textit{\textbf{mAP\textsubscript{0.5}}}} & \multicolumn{1}{l|}{\textit{\textbf{mAP\textsubscript{.5-.95}}}} & \multicolumn{1}{l|}{\textbf{\textit{F1-score}}} \\ \hline
\multicolumn{1}{|l|}{Bus}              & \multicolumn{1}{l|}{0.9331}          & \multicolumn{1}{l|}{0.4796}          & \multicolumn{1}{l|}{0.7156}                   & \multicolumn{1}{l|}{0.5419}                      & 0.6335                                 \\ \hline
\multicolumn{1}{|l|}{Bike}             & \multicolumn{1}{l|}{0.8035}          & \multicolumn{1}{l|}{0.377}           & \multicolumn{1}{l|}{0.6062}                   & \multicolumn{1}{l|}{0.3208}                      & 0.5132                                 \\ \hline
\multicolumn{1}{|l|}{Car}              & \multicolumn{1}{l|}{0.9493}          & \multicolumn{1}{l|}{0.5331}          & \multicolumn{1}{l|}{0.749}                    & \multicolumn{1}{l|}{0.5208}                      & 0.6827                                 \\ \hline
\multicolumn{1}{|l|}{Pedestrian}       & \multicolumn{1}{l|}{0.7785}          & \multicolumn{1}{l|}{0.1402}          & \multicolumn{1}{l|}{0.4596}                   & \multicolumn{1}{l|}{0.2168}                      & 0.2376                                 \\ \hline
\multicolumn{1}{|l|}{Truck}            & \multicolumn{1}{l|}{0.7444}          & \multicolumn{1}{l|}{0.3028}          & \multicolumn{1}{l|}{0.5424}                   & \multicolumn{1}{l|}{0.4141}                      & 0.4304                                 \\ \hline
\multicolumn{1}{|l|}{\textbf{All}}     & \multicolumn{1}{l|}{\textbf{0.8418}} & \multicolumn{1}{l|}{\textbf{0.3665}} & \multicolumn{1}{l|}{\textbf{0.6146}}          & \multicolumn{1}{l|}{\textbf{0.4029}}             & \textbf{0.5106}                        \\ \hline
\end{tabular}
\vspace{-2mm}
}
\caption{Results of YOLOv8x model on the input size 640 $\times$ 640.}
\label{table:4}
\vspace{-4mm}
\end{table}

\subsubsection{Results on Input Size 1280 $\times$ 1280}
\label{4:4:3}

Table \ref{table:2} shows that for an input size of 1280 $\times$ 1280, YOLOR-P6 \cite{YOLOR} and YOLOR-W6 \cite{YOLOR} achieved the highest $mAP\textsubscript{0.5}$ scores of 0.6632 and 0.6466, respectively. In contrast, YOLOv7-D6 \cite{wang2022yolov7} yielded the lowest $mAP\textsubscript{0.5}$ score of 0.3977. YOLOv7-E6E \cite{wang2022yolov7} demonstrated the highest performance in terms of \emph{F1-score} and \emph{Recall}, with values of 0.6294 and 0.5252, respectively.

\begin{figure}[t]
\centerline{
   \includegraphics[width=0.9\linewidth]{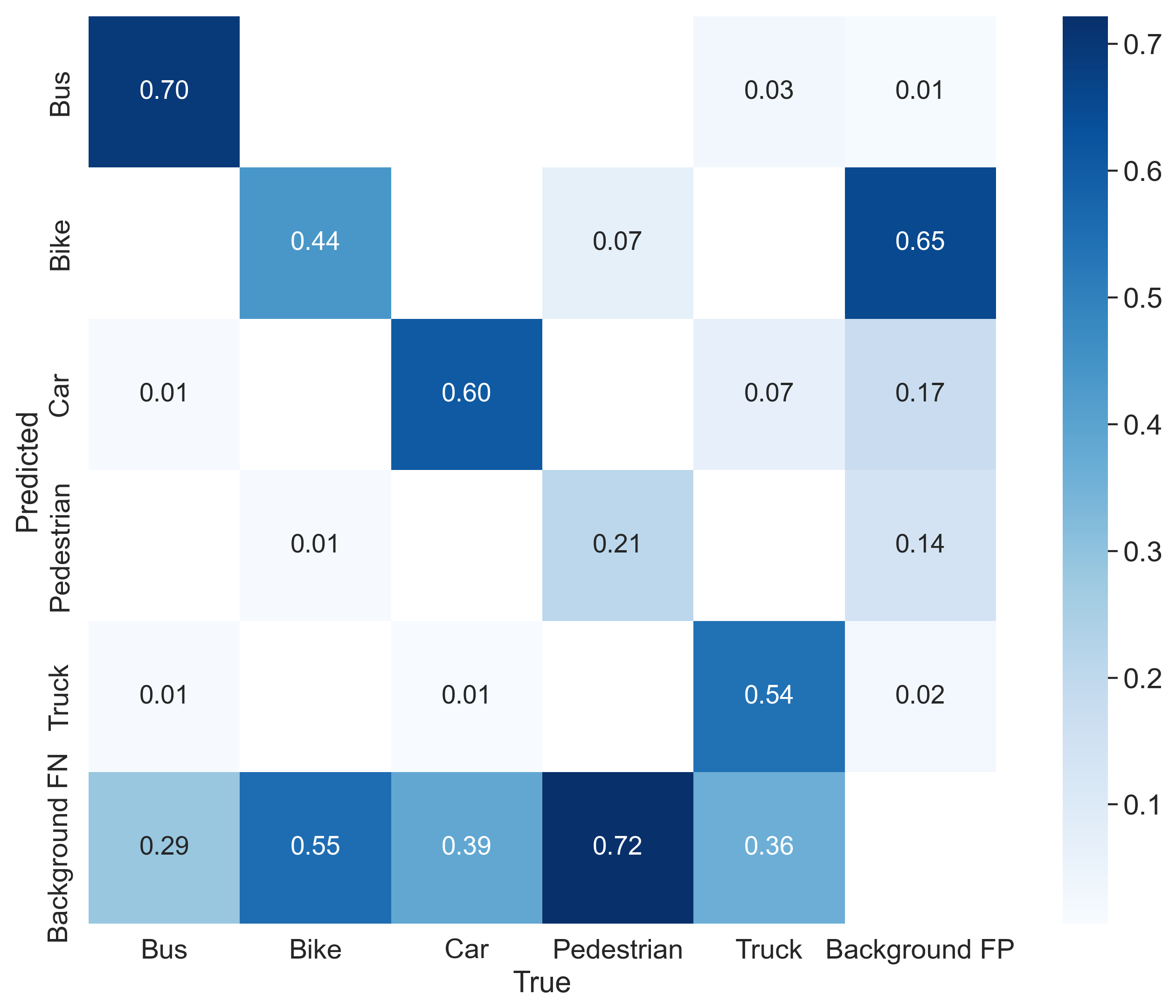}
\vspace{-2mm}
}   
\caption{Normalized Confusion Matrix of YOLOR-P6 model on input size 1280 $\times$ 1280.} 
\label{fig:confusionmatrix_yolor}
\vspace{-4mm}
\end{figure}

\begin{table}[t]
\footnotesize
\centerline{
\begin{tabular}{|llllll|}
\hline
\multicolumn{6}{|c|}{YOLOR-P6-1280$\times$1280 \cite{YOLOR}} \\ 
\hline
\multicolumn{1}{|l|}{\textbf{Classes}} & \multicolumn{1}{l|}{\textbf{\textit{Precision}}}      & \multicolumn{1}{l|}{\textbf{\textit{Recall}}}         & \multicolumn{1}{l|}{\textit{\textbf{mAP\textsubscript{0.5}}}} & \multicolumn{1}{l|}{\textit{\textbf{mAP\textsubscript{.5-.95}}}} & \multicolumn{1}{l|}{\textbf{\textit{F1-score}}} \\ \hline
\multicolumn{1}{|l|}{Bus}              & \multicolumn{1}{l|}{0.9429}          & \multicolumn{1}{l|}{0.6753}          & \multicolumn{1}{l|}{0.8161}                   & \multicolumn{1}{l|}{0.6271}                      & 0.7869                                \\ \hline
\multicolumn{1}{|l|}{Bike}             & \multicolumn{1}{l|}{0.8537}          & \multicolumn{1}{l|}{0.4316}          & \multicolumn{1}{l|}{0.6553}                   & \multicolumn{1}{l|}{0.3725}                      & 0.5733                                 \\ \hline
\multicolumn{1}{|l|}{Car}              & \multicolumn{1}{l|}{0.9473}          & \multicolumn{1}{l|}{0.6062}          & \multicolumn{1}{l|}{0.7876}                   & \multicolumn{1}{l|}{0.5575}                      & 0.7393                                 \\ \hline
\multicolumn{1}{|l|}{Pedestrian}       & \multicolumn{1}{l|}{0.4903}          & \multicolumn{1}{l|}{0.2014}          & \multicolumn{1}{l|}{0.3621}                   & \multicolumn{1}{l|}{0.2007}                      & 0.2855                                 \\ \hline
\multicolumn{1}{|l|}{Truck}            & \multicolumn{1}{l|}{0.7753}          & \multicolumn{1}{l|}{0.5541}           & \multicolumn{1}{l|}{0.695}                   & \multicolumn{1}{l|}{0.4451}                      & 0.6462                                 \\ \hline
\multicolumn{1}{|l|}{\textbf{All}}     & \multicolumn{1}{l|}{\textbf{0.8019}} & \multicolumn{1}{l|}{\textbf{0.4937}} & \multicolumn{1}{l|}{\textbf{0.6632}}          & \multicolumn{1}{l|}{\textbf{0.4406}}             & \textbf{0.6111}                        \\ \hline
\end{tabular}
\vspace{-2mm}
}
\caption{Results of YOLOR-P6 model on the input size 1280$\times$1280.}
\label{table:3}
\vspace{-4mm}
\end{table}

Furthermore, with regard to object scale, YOLOv7-E6E \cite{wang2022yolov7} exhibited higher performance over the other models in detecting small and medium-sized objects, achieving \emph{AP}s of 0.1684 and 0.5019, respectively. In contrast, YOLOR-P6 \cite{YOLOR} demonstrated exceptional accuracy in detecting large objects, with an \emph{AP\textsubscript{L}} of 0.7216.

Figure~\ref{fig:confusionmatrix_yolor} shows the confusion matrix and Table \ref{table:3} tabulates the results provided by the best-performing model YOLOR-P6 \cite{YOLOR} on the input size of 1280$\times$1280. The most accurately predicted class is Bus with an $mAP\textsubscript{0.5}$ of 0.8161 followed by Car, Truck, Bike and finally Pedestrian with $mAP\textsubscript{0.5}$ of 0.3621. 

The Bike has the maximum normalized $FP$ rate at 0.65 when the background is incorrectly detected as Bike. Additionally, a substantial fraction of objects in each class remains undetected, as indicated by their normalized $FN$ rates varying between 0.29 to 0.72. Despite this, the model demonstrates comparatively good performance in terms of $Precision$ across all classes, with values ranging from 0.77 to 0.95, with the exception of the Pedestrian class, which displays a significantly low $Precision$ of 0.49.

\subsubsection{Inference Time}
\label{4:4:4}
The inference time for each model was measured on a workstation featuring an 11$^{th}$ Gen i7 CPU and an Nvidia RTX 3080 GPU, and the results are presented in Table \ref{table:2}. The outcomes demonstrate that all models perform efficiently on this high-end computer, with inference times varying between 4.3 ms to 43.9 ms.

\section{Discussions}

The majority of the dataset, consisting of images from Cameras 1-15, were derived from fisheye surveillance camera footage captured on a single day in July 2018 in Taiwan. Although the dataset contains images of 5 major road participants captured from varying angles and under different illumination conditions, it lacks diversity in terms of weather conditions, such as fog, rain, snow, and storms. Additionally, the dataset is imbalanced, with the class Bike having the highest number of objects at 88K, while the Bus class has the lowest number at 2.98K.

\begin{figure}[t]
\centerline{
   \includegraphics[width=0.95\linewidth]{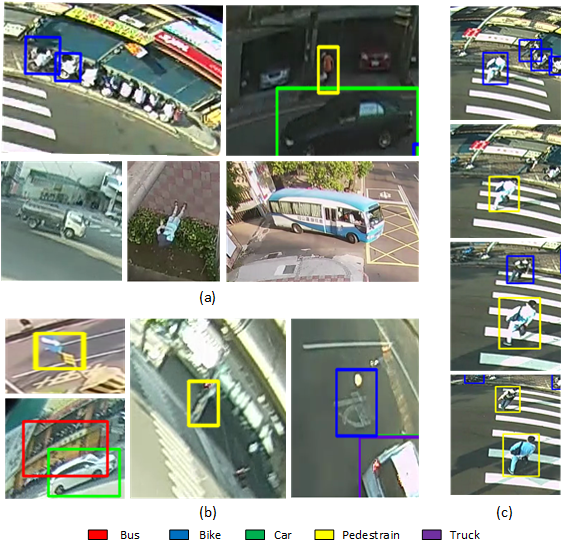}
\vspace{-2mm}
}   
\caption{Some samples of hard cases of YOLOR-P6 detections on input size 1280 $\times$ 1280.} 
\label{fig:hardcases}
\vspace{-4mm}
\end{figure}

\textbf{Hard cases} of the best-performing YOLOR-W6 \cite{YOLOR} are represented by few samples in Figure~\ref{fig:hardcases}. 

In Figure~\ref{fig:hardcases}(a), several examples of false negatives are shown where the labeled objects are not detected. These instances can be categorized into two groups: parked/stationary vehicles and road participants in motion. In the top left, only two out of nine scooters parked in a row on the sidewalk are correctly detected. On the top right, two partially visible cars parked in a garage are not detected. The presence of numerous parked vehicles in the dataset and the misdetection of such vehicles contribute to the high false negative rates observed across all classes. 

The second type of false positives involves road participants in motion, such as a truck, a pedestrian, and a bus shown in the three crops at the bottom of Figure~\ref{fig:hardcases}(a)

The examples shown in Figure~\ref{fig:hardcases}(b) illustrate instances where the background is misclassified as one of the object classes, resulting in higher false positive rates. Specifically, in the top left, a road sign is incorrectly detected as a Pedestrian, while in the bottom left, a yellow building is misclassified as a Bus. In the center, a building pillar is erroneously labeled as a Pedestrian, and on the right, a horizontal road sign is detected as a Bike.

In Figure~\ref{fig:hardcases}(c), we can observe cases where classes are misclassified as other classes. The four images, from the bottom to the top, show how the predictions change as Pedestrians walk away from the camera. We can see that misclassification occurs when the size of the objects gets smaller. Specifically, the objects were initially correctly detected as Pedestrians when they were closer to the camera, but as they moved away and became smaller, they were misclassified as Bikes.


\section{Conclusions}
\label{sec:conclusion}

We presented the FishEye8K benchmark dataset along with the evaluation of the SoTA one-stage object detectors for the use of fisheye cameras for road object detection. This dataset fills the gap in the lack of a fisheye surveillance camera dataset for road 2D object detection tasks. The anonymized dataset includes 8000 frames with 157K bounding boxes of 5 different road participants and various aspects of road conditions. Our evaluation results show that YOLOv8 and YOLOR models ~\cite{YOLOR}, which are pretrained on MS-COCO~\cite{COCO}, outperforms the other models. Therefore the FishEye8K dataset will be a significant contribution to the fisheye video analytics and smart city applications. 

{\bf Future work} includes the creation of a large and more balanced dataset with more diverse street object categories that can be used for object re-identification model training and evaluation.

{\bf Acknowledgements.}
Emirates Center for Mobility Research (EMCR) provided support for our research through Grant 12R012, while SciDM and National Center for High-performance Computing (NCHC) provided necessary storage resources. In addition, we thank AI \& Robotics Lab at the United Arab Emirates University for offering a DGX-1 GPU supercomputer.
{\small
\bibliographystyle{ieee_fullname}
\bibliography{egbib}

\begin{thebibliography}{10}\itemsep=-1pt

\bibitem{AAURainSnow}
Chris~H Bahnsen and Thomas~B Moeslund.
\newblock Rain removal in traffic surveillance: Does it matter?
\newblock {\em IEEE Transactions on Intelligent Transportation Systems},
  20(8):2802--2819, 2018.

\bibitem{UAVDT}
Dawei Du, Yuankai Qi, Hongyang Yu, Yifan Yang, Kaiwen Duan, Guorong Li, Weigang
  Zhang, Qingming Huang, and Qi Tian.
\newblock The unmanned aerial vehicle benchmark: Object detection and tracking.
\newblock In {\em ECCV}, pages 370--386, 2018.

\bibitem{Pascal-Voc}
M. Everingham, L. Van~Gool, C.~K.~I. Williams, J. Winn, and A. Zisserman.
\newblock The {PASCAL} {V}isual {O}bject {C}lasses {C}hallenge 2012 {(VOC2012)}
  {R}esults.
\newblock
  http://www.pascal-network.org/challenges/VOC/voc2012/workshop/index.html.

\bibitem{KITTI-D}
Andreas Geiger, Philip Lenz, and Raquel Urtasun.
\newblock Are we ready for autonomous driving? the kitti vision benchmark
  suite.
\newblock In {\em CVPR}, pages 3354--3361. IEEE, 2012.

\bibitem{DGX-1}
Tetiana Habuza, Khaled Khalil, Nazar Zaki, Fady Alnajjar, and Munkhjargal
  Gochoo.
\newblock Web-based multi-user concurrent job scheduling system on the shared
  computing resource objects.
\newblock In {\em 14th International Conference on Innovations in Information
  Technology (IIT)}. IEEE, Nov 2020.

\bibitem{CARPK}
Meng-Ru Hsieh, Yen-Liang Lin, and Winston~H Hsu.
\newblock Drone-based object counting by spatially regularized regional
  proposal network.
\newblock In {\em ICCV}, pages 4145--4153, 2017.

\bibitem{NEXET}
{Itay Klein, Nexar Blog}.
\newblock {NEXET} - the largest and most diverse road dataset in the world,
  2017.
\newblock
  \footnotesize{\url{https://data.getnexar.com/blog/nexet-the-largest-and-most-diverse-road-dataset-in-the-world}},
  Last accessed on 2021-10-24.

\bibitem{YOLOv5}
Glenn Jocher, Alex Stoken, Jirka Borovec, NanoCode012, ChristopherSTAN, Liu
  Changyu, Laughing, tkianai, yxNONG, Adam Hogan, lorenzomammana, AlexWang1900,
  Ayush Chaurasia, Laurentiu Diaconu, Marc, wanghaoyang0106, ml5ah, Doug,
  Durgesh, Francisco Ingham, Frederik, Guilhen, Adrien Colmagro, Hu Ye,
  Jacobsolawetz, Jake Poznanski, Jiacong Fang, Junghoon Kim, Khiem Doan, and
  Lijun Yu.
\newblock {ultralytics/yolov5: v4.0 - nn.SiLU() activations, Weights \& Biases
  logging, PyTorch Hub integration}, Jan. 2021.

\bibitem{ADAM}
Diederik~P. Kingma and Jimmy Ba.
\newblock Adam: A method for stochastic optimization, 2014.

\bibitem{Detection_in_LLC}
Roman Kvyetnyy, Roman Maslii, Volodymyr Harmash, Ilona Bogach, Andrzej Kotyra,
  aklin, Aizhan Zhanpeisova, and Nursanat Askarova.
\newblock Object detection in images with low light condition.
\newblock In {\em Photonics Applications in Astronomy, Communications,
  Industry, and High Energy Physics Experiments 2017}, volume 10445, page
  104450W. International Society for Optics and Photonics, 2017.

\bibitem{Focal}
Tsung{-}Yi Lin, Priya Goyal, Ross~B. Girshick, Kaiming He, and Piotr
  Doll{\'{a}}r.
\newblock Focal loss for dense object detection.
\newblock {\em CoRR}, abs/1708.02002, 2017.

\bibitem{COCO}
Tsung{-}Yi Lin, Michael Maire, Serge~J. Belongie, Lubomir~D. Bourdev, Ross~B.
  Girshick, James Hays, Pietro Perona, Deva Ramanan, Piotr Doll{\'{a}}r, and
  C.~Lawrence Zitnick.
\newblock Microsoft {COCO:} common objects in context.
\newblock {\em CoRR}, abs/1405.0312, 2014.

\bibitem{MIO-TCD}
Zhiming Luo, Frederic Branchaud-Charron, Carl Lemaire, Janusz Konrad, Shaozi
  Li, Akshaya Mishra, Andrew Achkar, Justin Eichel, and Pierre-Marc Jodoin.
\newblock Mio-tcd: A new benchmark dataset for vehicle classification and
  localization.
\newblock {\em IEEE Transactions on Image Processing}, 27(10):5129--5141, 2018.

\bibitem{paddle}
Yanjun Ma, Dianhai Yu, Tian Wu, and Haifeng Wang.
\newblock {PaddlePaddle}: An open-source deep learning platform from industrial
  practice.
\newblock {\em Frontiers of Data and Domputing}, 1(1):105, 2019.

\bibitem{AI-City}
Milind Naphade, Shuo Wang, David~C Anastasiu, Zheng Tang, Ming-Ching Chang,
  Xiaodong Yang, Yue Yao, Liang Zheng, Pranamesh Chakraborty, Christian~E
  Lopez, et~al.
\newblock The 5th ai city challenge.
\newblock In {\em CVPR}, pages 4263--4273, 2021.

\bibitem{MIT-Car}
Constantine Papageorgiou and Tomaso Poggio.
\newblock A trainable system for object detection.
\newblock {\em International journal of computer vision}, 38(1):15--33, 2000.

\bibitem{pytorch}
Adam Paszke, Sam Gross, Francisco Massa, Adam Lerer, James Bradbury, Gregory
  Chanan, Trevor Killeen, Zeming Lin, Natalia Gimelshein, Luca Antiga, Alban
  Desmaison, Andreas Kopf, Edward Yang, Zachary DeVito, Martin Raison, Alykhan
  Tejani, Sasank Chilamkurthy, Benoit Steiner, Lu Fang, Junjie Bai, and Soumith
  Chintala.
\newblock Pytorch: An imperative style, high-performance deep learning library.
\newblock In {\em Advances in Neural Information Processing Systems 32}, pages
  8024--8035. Curran Associates, Inc., 2019.

\bibitem{darknet}
Joseph Redmon.
\newblock Darknet: Open source neural networks in {C}.
\newblock \footnotesize{\url{http://pjreddie.com/darknet/}}, 2013--2016.

\bibitem{YOLO}
Joseph Redmon, Santosh~Kumar Divvala, Ross~B. Girshick, and Ali Farhadi.
\newblock You only look once: Unified, real-time object detection.
\newblock {\em CoRR}, abs/1506.02640, 2015.

\bibitem{YOLOR}
Chien{-}Yao Wang, I{-}Hau Yeh, and Hong{-}Yuan~Mark Liao.
\newblock You only learn one representation: Unified network for multiple
  tasks.
\newblock {\em CoRR}, abs/2105.04206, 2021.

\bibitem{wang2022yolov7}
Chien-Yao Wang, Alexey Bochkovskiy, and Hong-Yuan~Mark Liao.
\newblock {YOLOv7}: Trainable bag-of-freebies sets new state-of-the-art for
  real-time object detectors.
\newblock {\em arXiv preprint arXiv:2207.02696}, 2022.

\bibitem{UA-DETRAC}
Longyin Wen, Dawei Du, Zhaowei Cai, Zhen Lei, Ming{-}Ching Chang, Honggang Qi,
  Jongwoo Lim, Ming{-}Hsuan Yang, and Siwei Lyu.
\newblock {DETRAC:} {A} new benchmark and protocol for multi-object tracking.
\newblock {\em CoRR}, abs/1511.04136, 2015.

\bibitem{BDD100K}
Huazhe Xu, Yang Gao, Fisher Yu, and Trevor Darrell.
\newblock End-to-end learning of driving models from large-scale video
  datasets.
\newblock In {\em CVPR}, pages 2174--2182, 2017.

\bibitem{woodscape}
Senthil Yogamani, Ciar{\'a}n Hughes, Jonathan Horgan, Ganesh Sistu, Padraig
  Varley, Derek O'Dea, Michal Uric{\'a}r, Stefan Milz, Martin Simon, Karl
  Amende, et~al.
\newblock {WoodScape}: A multi-task, multi-camera fisheye dataset for
  autonomous driving.
\newblock In {\em ICCV}, pages 9308--9318, 2019.

\bibitem{yu2020bdd100k}
Fisher Yu, Haofeng Chen, Xin Wang, Wenqi Xian, Yingying Chen, Fangchen Liu,
  Vashisht Madhavan, and Trevor Darrell.
\newblock Bdd100k: A diverse driving dataset for heterogeneous multitask
  learning.
\newblock In {\em CVPR}, pages 2636--2645, 2020.

\end{thebibliography}
}
\end{document}